\pdfoutput=1
\documentclass{article}
\usepackage{spconf,amsmath,amssymb,graphicx,booktabs,tabularx,hyperref}
\usepackage[table]{xcolor}
\usepackage{tikz}
\renewcommand{\paragraph}[1]{\par\addvspace{1.1ex plus .3ex minus .2ex}\noindent\textbf{#1}\hspace{.6em}\ignorespaces}
\usetikzlibrary{arrows.meta,positioning,calc,shapes.geometric}
\hypersetup{colorlinks=false,pdfborder={0 0 0}}

\def\x{{\mathbf x}}

\title{SCALE-RECURSIVE RECTIFIED FLOWS FOR FEW-STEP PRECIPITATION ENSEMBLES}

\name{Shunya Nagashima$^{1\ast}$ \qquad Takumi Bannai$^{2,3\ast}$%
  \thanks{$^{\ast}$Equal contribution.}}
\address{$^{1}$Neurogica Inc.\quad $^{2}$LTS, Inc.\quad $^{3}$ME-Lab Japan, Inc.}

\newcommand{\EBatchFlatSixteen}{3.72}
\newcommand{\EBatchAllocated}{1.84}

\newcommand{\ESpeedRange}{1.8--2.0}

\begin{document}
\brokenpenalty=10000 %
\ninept
\maketitle

\begin{abstract}
\hspace*{\parindent}%
Fine-resolution precipitation estimates support flood risk assessment and water management,
but coarse satellite products cannot resolve rainfall within each grid cell.
Generative models address this ambiguity by producing ensembles of plausible high-resolution
rainfall fields. Among these models, rectified flows generate samples by iteratively
transforming random noise into rainfall fields. Reducing the number of sampling steps accelerates generation but can
make ensemble members too similar, understating uncertainty.
We propose a scale-recursive rectified flow that generates broad patterns before local details
and guides sampling-step allocation by comparing ensemble variability with prediction error
across spatial scales. Validation scores and rainfall power spectra constrain the allocation
to avoid excessive amplification.
In satellite-to-radar downscaling over the contiguous United States, our analysis identified
broad rainfall patterns as the main source of insufficient ensemble variability under reduced
sampling budgets. Allocating more steps to the coarse flow improved probabilistic accuracy
and rain detection across training seeds at fixed architecture and computational cost.
The proposed model also achieved better probabilistic accuracy with shorter sampling time
than a nonrecursive flow using more steps.
\end{abstract}


\section{Introduction}
\label{sec:intro}

Fine-resolution rainfall estimates support flood risk assessment and water management
\cite{ravuri2021skilful,harris2022stochastic}. Satellite products extend coverage beyond
radar networks, but combine observations with limited spatial and temporal resolution.
IMERG merges microwave and infrared estimates onto a $0.1^\circ$ grid
\cite{huffman2020imerg,nasa2023imerg}, so a cell of roughly ten kilometers cannot specify
the distribution of rain at kilometer scales.

Different spatial distributions of rainfall can produce the same coarse-grid average.
Generative downscaling represents this ambiguity by sampling multiple plausible
high-resolution rainfall maps
\cite{harris2022stochastic,mardani2025corrdiff}. We call insufficient ensemble variability
relative to prediction error a \emph{spread deficit}, following the spread--error relation
\cite{fortin2014spread}. Flow-based generative models, including rectified flows
\cite{liu2023rectified,lipman2023flow}, generate samples by numerically integrating learned
trajectories from noise to data. Each integration step requires a neural-network evaluation;
fewer steps save computation but can increase this deficit.

Existing methods improve sampling through trajectory straightening \cite{liu2023rectified},
wavelet hierarchies with equal step counts across scales \cite{guth2022wsgm}, or noise and
interpolation schedules guided by spectra \cite{chen2025scaleadaptive}.
These designs do not determine how a fixed budget should be divided between resolution
levels from conditional ensemble spread relative to prediction error. We address this gap
by diagnosing the spread deficit at each scale and using its response to sampling steps
to predict an allocation. Fixed-model, equal-cost controls then test the predicted benefit. Our contributions are
as follows.
\begin{itemize}
\setlength{\itemsep}{0pt}\setlength{\parsep}{0pt}
\item We measure spread relative to prediction error in wavelet bands and identify broad
rainfall patterns as the main source of the few-step spread deficit.
\item We predict a coarse-heavy allocation from the measured deficit, then test it at fixed
sampling cost with spectral checks for excessive amplification.
\item We establish allocation gains with fixed-model, equal-cost controls across training
seeds, measure sampling speed, and quantify the heavy-rain tradeoffs of optional calibration.
\end{itemize}

\begin{figure*}[t]
\setlength{\abovecaptionskip}{4pt}
\centering
\resizebox{\textwidth}{!}{%
\begin{tikzpicture}[x=.975mm,y=.7mm,font=\scriptsize,>={Stealth[length=1.5mm]},
  every node/.style={align=center,inner sep=2pt},
  imagebox/.style={inner sep=0pt,draw=black!40},
  process/.style={draw=black!55,fill=black!4,rounded corners=1pt,minimum height=4mm,inner sep=1pt},
  flow/.style={process,draw=blue!55!black,fill=blue!7,minimum width=23mm,minimum height=7mm},
  data/.style={draw=black!35,fill=white,rounded corners=1pt,minimum height=4mm,inner sep=1pt},
  path/.style={->,line width=.6pt,black!70},
  condition/.style={->,dashed,line width=.6pt,black!65},
  sample/.style={path,blue!65!black},
  train/.style={->,>={Stealth[length=1.15mm]},line width=.65pt,orange!80!black},
  traincond/.style={train,dashed},
  target/.style={data,draw=orange!80!black,fill=orange!7},
  dwtblock/.style={minimum width=7mm,minimum height=4mm,inner sep=1pt},
  panel/.style={draw=black!45,fill=black!2,line width=.4pt},
  lab/.style={font=\scriptsize,text=black!75},
  img/.style={inner sep=0pt,draw=black!35}]
  \node[imagebox] (u) at (1,3) {\includegraphics[width=8mm]{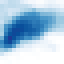}};
  \node[below=.3mm of u] {LR input $u$};
  \node[process] (predict) at (25,3) {Predictor};
  \node[imagebox] (s) at (48,3) {\includegraphics[width=8mm]{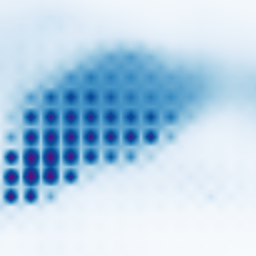}};
  \node[below=.3mm of s] {prediction $s$};
  \node[process,dwtblock] (dwt) at (76,3) {DWT};
  \node[process] (cc) at (76,-8) {Low-pass};
  \node[process,dwtblock] (cdwt) at (61,-8) {DWT};
  \node[data] (cf) at (111,3) {$\mathcal W(s)$};
  \draw[path] (u) -- (predict); \draw[path] (predict) -- (s);
  \draw[path] (s) -- (dwt);
  \draw[path] (dwt.south) -- (cc.north);
  \draw[path] (dwt.east) -- node[above=.4mm,inner sep=.5pt] {all bands} (cf.west);
  \draw[path] (cc.west) -- node[above=.5mm,inner sep=.5pt] {$L_s$} (cdwt.east);
  \node[imagebox] (noise_c) at (1,-22) {\includegraphics[width=6mm]{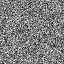}};
  \node[above=.1mm of noise_c] {noise $z_c$};
  \node[flow] (coarse) at (25,-22) {Coarse flow $v_c$\\$N_c$ Euler steps};
  \node[process] (inv_c) at (70,-22) {IDWT};
  \node[data] (ll) at (89,-22) {low-pass $L$};
  \draw[sample] (noise_c) -- (coarse);
  \draw[sample] (coarse.east) -- node[above=.4mm,inner sep=.5pt] {$\mathcal W(L)$} (inv_c.west);
  \draw[sample] (inv_c) -- (ll);
  \draw[condition,rounded corners=.8mm] (cdwt.south) -- (61,-13) -- node[above=.5mm,pos=.65,inner sep=.5pt] {$\mathcal W(L_s)$} (25,-13) -- (coarse.north);
  \node[imagebox] (noise_f) at (50,-41) {\includegraphics[width=6mm]{assets/noise.png}};
  \node[above=.6mm of noise_f,xshift=4mm] {noise $z_f$};
  \node[flow] (fine) at (77,-41) {Fine flow $v_f$\\$N_f$ Euler steps};
  \draw[sample] (noise_f) -- (fine);
  \draw[condition,rounded corners=.8mm] (cf.south) -- (111,-32) -- (83,-32) -- ([xshift=5.85mm]fine.north);
  \draw[condition,blue!65!black,rounded corners=.8mm] (ll.south) -- (89,-29) -- (71,-29) -- ([xshift=-5.85mm]fine.north);
  \node[process] (inv_r) at (116,-41) {IDWT};
  \node[below=.3mm of inv_r] {$[L;D]\mapsto r$};
  \draw[sample] (fine.east) -- node[above] {details $D$} (inv_r.west);
  \draw[sample,preaction={draw=white,line width=2.5pt}] (ll.east) -- (116,-22) -- (inv_r.north);
  \node[process,circle,minimum size=3.5mm,inner sep=0pt] (add) at (138,-41) {$+$};
  \draw[sample] (inv_r) -- node[above] {$r$} (add);
  \draw[path,rounded corners=1mm] (s.north) -- (48,12) -- (123,12) -- (123,-36) -- (138,-36) -- node[right] {$s$} (add.north);
  \node[process] (units) at (155,-41) {$T^{-1}$};
  \draw[sample] (add) -- node[above] {$s+r$} (units);
  \node[imagebox] (out) at (173,-41) {\includegraphics[width=8mm]{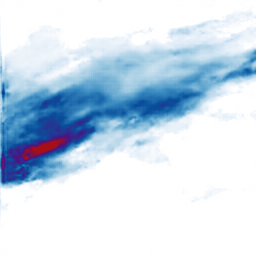}};
  \draw[sample] (units) -- (out);
  \node[below=1mm of out] {$\hat y^{(m)}$};
  \node[imagebox,draw=orange!80!black] (truth) at (7,-66)
    {\includegraphics[width=6mm]{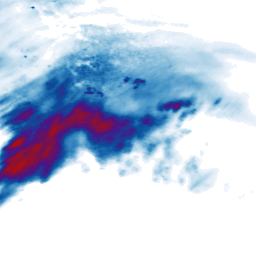}};
  \node[anchor=west] at (13,-67) {HR target $y$};
  \node[imagebox,draw=orange!80!black] (trres) at (7,-54) {\includegraphics[width=6mm]{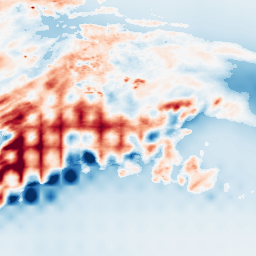}};
  \node[left=.7mm of trres] {$T(y)-s$};
  \node[target,dwtblock] (trbands) at (7,-43) {DWT};
  \node[target] (lgt) at (42,-43) {$L_{\rm gt}$};
  \node[target,minimum height=4mm] (dgt) at (25,-57) {$D_{\rm gt}$};
  \draw[train] (truth) -- (trres);
  \draw[train] (trres) -- (trbands);
  \draw[train] (trbands.east) -- node[above=0mm,xshift=-3mm,inner sep=.5pt,text=black] {$\mathcal W(T(y)-s)$} (lgt.west);
  \draw[train] (25,-43) -- (dgt.north);
  \node[target,circle,minimum size=2.4mm,inner sep=0pt] (interp_c) at (25,-33.5) {$+$};
  \node[target,dwtblock] (target_dwt) at (42,-33.5) {DWT};
  \draw[train,rounded corners=.8mm] (noise_c.south) -- (1,-33.5) -- (interp_c.west);
  \draw[train] (lgt.north) -- (target_dwt.south);
  \draw[train] (target_dwt.west) -- node[above=.3mm,inner sep=.5pt,text=black] {$\mathcal W(L_{\rm gt})$} (interp_c.east);
  \draw[train] (interp_c.north) -- (coarse.south);
  \draw[traincond,rounded corners=.8mm] (lgt.south) -- (42,-51)
    -- node[above,text=black] {$+\,\epsilon$} (71,-51) -- ([xshift=-5.85mm]fine.south);
  \node[target,circle,minimum size=2.4mm,inner sep=0pt] (interp_f) at (83,-57) {$+$};
  \draw[train,preaction={draw=white,line width=2.5pt}] (noise_f.south)
    -- (50,-57) -- (interp_f.west);
  \draw[train,rounded corners=1mm] (dgt.south) -- (25,-63) -- (83,-63) -- (interp_f.south);
  \draw[train] (interp_f.north) -- ([xshift=5.85mm]fine.south);
  \draw[path] (113,-54) -- (120,-54);
  \node[anchor=west] at (121,-54) {shared};
  \draw[sample] (143,-54) -- (150,-54);
  \node[anchor=west] at (151,-54) {inference};
  \draw[train] (113,-60) -- (120,-60);
  \node[anchor=west] at (121,-60) {training only};
  \draw[condition] (143,-60) -- (150,-60);
  \node[anchor=west] at (151,-60) {condition};
  \begin{scope}[shift={(127mm,-21mm)}]
    \node[panel, minimum width=44mm, minimum height=33mm, anchor=south west] (pn) at (0,-3mm) {};
    \node[lab, anchor=north west] at ($(pn.north west)+(1mm,-0.5mm)$) {\textbf{Rectified flow sampling}};
    \tikzset{dens/.style={draw=blue!55!black, line width=0.45pt, shade, shading=axis, shading angle=90, left color=blue!30, right color=blue!3}}
    \foreach \xc in {6,22,38}{\draw[black!35, line width=0.3pt] (\xc mm,14.2mm) -- (\xc mm,24.6mm);}
    \path[dens] (6mm,14.55mm) -- plot[domain=-3:3, samples=70, smooth] ({6mm + 5.2mm*exp(-\x*\x/2)}, {19.2mm + 1.55mm*\x}) -- (6mm,23.85mm) -- cycle;
    \path[dens] (22mm,14.55mm) -- plot[domain=-3:3, samples=70, smooth] ({22mm + 5.2mm*((\x+3)^2.2)*exp(-(\x+3)/0.75)/(1.65^2.2*exp(-2.2))}, {19.2mm + 1.55mm*\x}) -- (22mm,23.85mm) -- cycle;
    \path[dens] (38mm,14.55mm) -- plot[domain=-3:3, samples=90, smooth] ({38mm + 5.2mm*((\x+3)^0.5)*exp(-(\x+3)/0.7)/(0.35^0.5*exp(-0.5))}, {19.2mm + 1.55mm*\x}) -- (38mm,23.85mm) -- cycle;
    \node[lab, anchor=south] at (8.5mm,24.4mm) {$p_0$}; \node[lab, anchor=south] at (24.5mm,24.4mm) {$p_t$}; \node[lab, anchor=south] at (40.5mm,24.4mm) {$p_1$};
    \newcommand{\traj}[5]{%
      \foreach \i in {0,...,23}{
        \pgfmathsetmacro{\h}{\i<12 ? 0 : 1}
        \pgfmathsetmacro{\sa}{(\i-12*\h)/12}\pgfmathsetmacro{\sb}{(\i+1-12*\h)/12}
        \pgfmathsetmacro{\ya}{\h==0 ? #1 : #2}\pgfmathsetmacro{\yb}{\h==0 ? #2 : #3}
        \pgfmathsetmacro{\xo}{6+16*\h}
        \pgfmathsetmacro{\Xa}{\xo + 16*(1.5*(1-\sa)*(1-\sa)*\sa + 1.5*(1-\sa)*\sa*\sa + \sa*\sa*\sa)}
        \pgfmathsetmacro{\Xb}{\xo + 16*(1.5*(1-\sb)*(1-\sb)*\sb + 1.5*(1-\sb)*\sb*\sb + \sb*\sb*\sb)}
        \pgfmathsetmacro{\Ya}{19.2 + 1.55*(\ya + (\yb-\ya)*(3*\sa*\sa - 2*\sa*\sa*\sa))}
        \pgfmathsetmacro{\Yb}{19.2 + 1.55*(\ya + (\yb-\ya)*(3*\sb*\sb - 2*\sb*\sb*\sb))}
        \pgfmathsetmacro{\pc}{100*\i/23}
        \draw[line width=#4, line cap=round, opacity=#5, color=blue!\pc!violet!80!black] (\Xa mm,\Ya mm) -- (\Xb mm,\Yb mm);}}
    \traj{-1.8}{-2.4}{-2.91}{0.35pt}{0.55}
    \traj{-1.1}{-1.95}{-2.78}{0.35pt}{0.55}
    \traj{-0.5}{-1.4}{-2.52}{0.35pt}{0.55}
    \traj{0.5}{-0.2}{-1.72}{0.35pt}{0.55}
    \traj{1.1}{0.6}{-1.05}{0.35pt}{0.55}
    \traj{1.8}{1.9}{0.0}{0.35pt}{0.55}
    \traj{0.0}{-0.85}{-2.17}{1.1pt}{1.0}
    \fill[violet!80!black] (6mm,19.2mm) circle (0.55mm); \fill[blue!50!violet!80!black] (22mm,{19.2mm-1.55mm*0.85}) circle (0.55mm); \fill[blue!80!black] (38mm,{19.2mm-1.55mm*2.17}) circle (0.55mm);
    \node[img] at (6mm,8.2mm) {\includegraphics[width=6mm]{assets/noise.png}};
    \node[img] at (22mm,8.2mm) {\includegraphics[width=6mm]{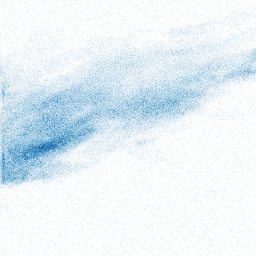}};
    \node[img] at (38mm,8.2mm) {\includegraphics[width=6mm]{assets/member0.png}};
    \node[lab] at (6mm,12.6mm) {$\x_0$}; \node[lab] at (22mm,12.6mm) {$\x_t$}; \node[lab] at (38mm,12.6mm) {$\x_1$};
    \draw[->, black!60, line width=0.5pt] (9.5mm,2.2mm) -- (34.5mm,2.2mm);
    \node[lab] at (5.5mm,2.2mm) {$t{=}0$}; \node[lab] at (38.5mm,2.2mm) {$t{=}1$};
    \node[lab, anchor=north] at (22mm,1.4mm) {Euler, $N$ steps};
  \end{scope}
\end{tikzpicture}}
\caption{LR input $u$ and HR target $y$. Blue: inference; orange: training; gray: shared.
Dashed arrows supply conditions; orange $\oplus$ denotes training interpolation (Sec.~\ref{sec:pyramid}).}
\label{fig:system}
\end{figure*}

\begin{figure}[t]
\centering
\includegraphics[width=\columnwidth]{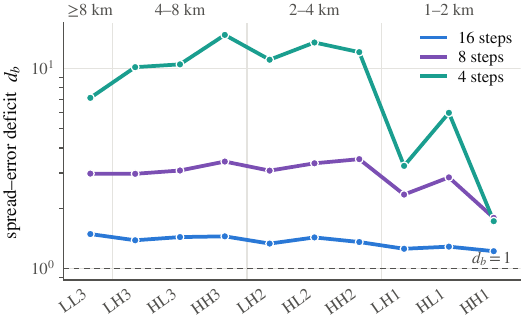}
\caption{Spread--error deficit of the single flow on 24 validation windows, by wavelet
band from coarse (LL3, wavelengths $\ge8$\,km) to fine (1--2\,km details). $d_b=1$ marks
aggregate spread--error balance, not full calibration.}
\label{fig:deficit}
\end{figure}

\section{Related work}
\label{sec:prior}

Harris et al.\ \cite{harris2022stochastic} use adversarial models for stochastic
precipitation downscaling. CorrDiff \cite{mardani2025corrdiff} adds diffusion-generated
residuals to a deterministic prediction, while NSP \cite{nagashima2026nsp} combines
gauges with spatiotemporal modeling for satellite precipitation refinement.
Closely related, Yi et al.\ \cite{yi2026wavelet} downscale precipitation with a wavelet diffusion model regularized by total variation on detail coefficients.

Cascaded diffusion \cite{ho2022cascaded} and WSGM \cite{guth2022wsgm} factor generation
across resolutions; WSGM uses equal step counts per scale. Xiao et al.\
\cite{xiao2024multiscale} combine low-frequency score models with high-frequency
adversarial generation. We hold the hierarchy fixed and vary only the step~allocation.

Reflow \cite{liu2023rectified} straightens trajectories; colored-noise sampling
\cite{cns2026colored} varies noise by frequency. Scale-adaptive flows
\cite{chen2025scaleadaptive} use spectral and non-Gaussianity diagnostics to design
noise and interpolation schedules. We instead use conditional spread--error measurements
to allocate integration steps. Variance inflation and EMOS \cite{gneiting2005emos}
adjust dispersion after prediction; our allocation precedes such correction and is checked with spatial spectra.

\section{System and diagnostic}
\label{sec:system}

\subsection{Conditional residual generation}

Following corrective diffusion \cite{mardani2025corrdiff}, we first predict a high-resolution
field and then add sampled residuals that represent the uncertainty around it. We write $u$ for the coarse input and $y$ for the high-resolution target, and map $y$ to the
modeling space with an invertible transform $T$, $z=T(y)$. The predictor
outputs $s=f_\phi(u)$ in this space, and we train the flows on the residual $r=z-s$.
Our single-flow baseline generates all residual bands jointly. At inference, it samples residual wavelet coefficients conditioned on
$\mathcal W(s)$, where $\mathcal W$ is the discrete wavelet transform (DWT). We apply the inverse transform (IDWT) to obtain a residual field $r^{(m)}$ for each ensemble
member $m$, add it to $s$, and apply $T^{-1}$ to recover the original units,
$\hat y^{(m)}=T^{-1}(s+r^{(m)})$.

\subsection{Spread--error diagnostic}

An aggregate spread statistic cannot identify which scales need more computation, so we
compare variability with prediction error in each wavelet band across sampling budgets. For band $b$ of $z$, we compute
\begin{equation}
d_b = \frac{\mathbb E[(\bar{x}_b-x_b^{\mathrm{obs}})^2]}
{(1+M^{-1})\,\mathbb E[\sigma_b^2]},
\label{eq:deficit}
\end{equation}
where $M$ denotes ensemble size, $\bar{x}_b$ and $\sigma_b^2$ are the ensemble mean and
unbiased sample variance at a coefficient in band $b$, and $x_b^{\mathrm{obs}}$ is its
observed value. We average the expectations over validation cases and valid coefficients.
The reference $d_b=1$ indicates aggregate spread--error balance \cite{fortin2014spread};
$d_b>1$ indicates insufficient spread. Excessive spread in light rain can offset insufficient
spread in heavy rain, so $d_b=1$ alone does not establish calibration.
Fig.~\ref{fig:deficit} shows $d_b$ for the single flow at 16, 8 and 4 steps. With sixteen
steps, all bands lie between 1.2 and 1.5. With fewer steps the deficit grows unevenly. At
four steps, the bands with wavelengths of 2\,km and longer reach 7--15, while the 1--2\,km
details stay between 1.7 and 6.0. The few-step deficit therefore sits in the broad rainfall
patterns. The model of Sec.~\ref{sec:pyramid} generates these bands with a separate coarse
flow whose steps are cheaper, so the diagnostic suggests spending steps there.

\section{Scale-recursive flow}
\label{sec:pyramid}

\subsection{Coarse and fine flows}

A single flow updates every wavelet band at each Euler step, so it cannot spend more steps
on the bands where Fig.~\ref{fig:deficit} finds the deficit. We therefore generate the
low-pass field and the details with two flows, one on the coarse grid and one on the fine
grid, and assign steps to each.
We define the ground-truth low-pass and detail bands as $[L_{\rm gt};D_{\rm gt}]=\mathcal W(T(y)-s)$
and use $x_1^c=\mathcal W(L_{\rm gt})$ and $x_1^f=D_{\rm gt}$ as the training endpoints of the
two flows, on grids with sides $h_c$ and $h_f$ (Fig.~\ref{fig:system}). The low-pass band lives on a smaller grid than the details, so a coarse step processes
fewer coefficients and costs less than a fine step. We take the cost of a step to be
proportional to its grid area (Sec.~\ref{sec:exp}).
We train each flow to predict the velocity $x_1-x_0$ from $x_t=(1-t)x_0+t x_1$, with
$x_0\sim\mathcal N(0,I)$ and $t\sim\mathcal U[0,1]$.
We condition the coarse flow on $\mathcal W(L_s)$, where $L_s$ is the low-pass band of
$\mathcal W(s)$, and the fine flow on $\mathcal W(s)$ and $L_{\rm gt}+\epsilon$ concatenated across
channels, with $\epsilon\sim\mathcal N(0,\sigma_{\mathrm{cond}}^2 I)$ \cite{ho2022cascaded}.
At inference, we draw independent Gaussian initial states $z_c,z_f$, generate $\mathcal W(L)$
and $D$, reconstruct $L$ by the inverse transform, and use it in place of $L_{\rm gt}+\epsilon$
in the fine condition; no target $y$ is used.

\subsection{Diagnostic-guided allocation}

With two flows, a sampling budget can be split in many ways, and the best split is not
known in advance. We choose it on validation data after training, with the model fixed,
so one trained model serves every budget. We write $N_c,N_f$ for the
coarse and fine step counts and $c_c,c_f$ for their per-step costs, and for a budget $B$ we require
\begin{equation}
 c_cN_c+c_fN_f\le B,\qquad N_c,N_f\in\mathbb N_{>0},
\label{eq:allocation}
\end{equation}
where equal-compute allocation splits the budget equally. On validation data we vary
step counts and compare the continuous ranked probability score (CRPS) and band-wise
deficits. If coarse steps help and fine steps do not, we include
$N_f=1$, $N_c=\lfloor(B-c_f)/c_c\rfloor$ as a coarse-heavy candidate.
We compare it with equal-compute and intermediate allocations, inspecting spectral power
alongside CRPS, and fix the selected steps without retraining.

\subsection{Spectral checks and calibration}
\label{sec:spectral}

Allocation reduces the coarse-scale deficit but may not remove it. Adding spread lowers
CRPS even when the added variation is unrealistic. We therefore compare the spatial power
of the members with the observed power, and we use this check to bound an optional
calibration of the low-pass band. For a wavelength range $\Omega$, $P_\Omega$ is the ratio of ensemble-averaged spectral power to observed power. A value of
one means equal power; values below one mean that the members are smoother than the
observations in that band, and values above one mean that they are rougher.

We optionally adjust the spread of the generated low-pass coefficients $L^{(m)}$.
With their ensemble mean $\bar L=M^{-1}\sum_{m=1}^{M}L^{(m)}$ and a shared gain $g>0$, we set
$L_{\mathrm{cal}}^{(m)}=\bar L+g(L^{(m)}-\bar L)$.
Thus $g>1$ expands member deviations, preserving their mean and the detail coefficients $D$.
We select the $g$ with the lowest validation CRPS under two constraints. First,
$P_\Omega\in[P_{\min},P_{\max}]$, where $P_{\min}$ and $P_{\max}$ set how much smoother or rougher
than the observations the members may be. Second, $d_b\ge1$ in every band
(Eq.~\ref{eq:deficit}), so that spread does not exceed aggregate error.
We fix the gain for testing.
For comparison, we also evaluate all-band spread scaling, which applies the validation-fitted gain $g_b=\sqrt{\max(d_b,1)}$ in each band $b$.

\section{Experiments}
\label{sec:exp}

\subsection{Experimental setup}

\begin{table}[t]
\centering\scriptsize
\caption{Matched-cost test scores by budget. Shaded rows are our allocations, named by
$(N_c,N_f)$, the coarse and fine step counts. Bold marks tier-best values; asterisks mark differences from the band-scaled single flow.
Italics: the sixteen-step single flow, listed in the cost-eight tier as the speed reference.}
\label{tab:cost}
\setlength{\tabcolsep}{1.25pt}
\renewcommand{\arraystretch}{1.08}
\setlength{\aboverulesep}{0pt}\setlength{\belowrulesep}{0pt}
\resizebox{\columnwidth}{!}{\begin{tabular}{lcccc}
\toprule
configuration & CRPS$\downarrow$ & CRPS$_{10}\downarrow$ & BSS$_{10}\uparrow$ & CSI$_{10}\uparrow$ \\
\midrule
\rowcolor{gray!14}\multicolumn{5}{@{}l}{\emph{cost 4 (single-flow Euler steps)}} \\
single flow & 0.0821$^{*}$ & 14.15$^{*}$ & +.004$^{*}$ & 0.003$^{*}$ \\
pixel-domain flow 600 h & 0.0865$^{*}$ & 13.81$^{*}$ & +.013$^{*}$ & 0.009$^{*}$ \\
single flow, reflowed \cite{liu2023rectified} & 0.0740$^{*}$ & 12.89$^{*}$ & +.038$^{*}$ & 0.032$^{*}$ \\
single flow, reflowed + band scaling & 0.0717$^{*}$ & 12.41$^{*}$ & +.040$^{*}$ & 0.040$^{*}$ \\
single flow + band scaling & 0.0726\phantom{$^{*}$} & 12.69\phantom{$^{*}$} & \textbf{+.046}\phantom{$^{*}$} & 0.037\phantom{$^{*}$} \\
\rowcolor{blue!9}\textbf{ours} (12,1) & 0.0727\phantom{$^{*}$} & 13.45$^{*}$ & +.031$^{*}$ & 0.025$^{*}$ \\
\rowcolor{blue!9}\textbf{ours} (12,1) + band scaling & \textbf{0.0688}$^{*}$ & \textbf{12.14}$^{*}$ & +.031$^{*}$ & \textbf{0.044}$^{*}$ \\
\rowcolor{blue!9}\textbf{ours} (12,1) + low-pass scaling & 0.0695$^{*}$ & 12.60$^{*}$ & +.042\phantom{$^{*}$} & 0.041$^{*}$ \\
\midrule
\rowcolor{gray!14}\multicolumn{5}{@{}l}{\emph{cost 8 (single-flow Euler steps)}} \\
single flow & 0.0760$^{*}$ & 13.75$^{*}$ & +.016$^{*}$ & 0.012$^{*}$ \\
pixel-domain flow 600 h & 0.0778$^{*}$ & 13.42$^{*}$ & +.026$^{*}$ & 0.020$^{*}$ \\
single flow, reflowed \cite{liu2023rectified} & 0.0738$^{*}$ & 12.89$^{*}$ & +.038$^{*}$ & 0.032$^{*}$ \\
single flow, reflowed + band scaling & 0.0716$^{*}$ & 12.43$^{*}$ & +.040\phantom{$^{*}$} & 0.040$^{*}$ \\
single flow + churn \cite{karras2022edm} & 0.0950$^{*}$ & 13.58$^{*}$ & +.020$^{*}$ & 0.015$^{*}$ \\
single flow + repulsion \cite{corso2024particle} & 0.0919$^{*}$ & 13.33$^{*}$ & +.029$^{*}$ & 0.022$^{*}$ \\
\emph{single flow, 16 steps} & \emph{0.0727\phantom{$^{*}$}} & \emph{13.08\phantom{$^{*}$}} & \emph{+.036\phantom{$^{*}$}} & \emph{0.030\phantom{$^{*}$}} \\
single flow + band scaling & 0.0708\phantom{$^{*}$} & 12.79\phantom{$^{*}$} & \textbf{+.041}\phantom{$^{*}$} & 0.034\phantom{$^{*}$} \\
equal compute (16,4) & 0.0715\phantom{$^{*}$} & 13.14$^{*}$ & +.039\phantom{$^{*}$} & 0.033\phantom{$^{*}$} \\
\rowcolor{blue!9}\textbf{ours} (28,1) & 0.0708\phantom{$^{*}$} & 12.59$^{*}$ & \textbf{+.041}\phantom{$^{*}$} & 0.044$^{*}$ \\
\rowcolor{blue!9}\textbf{ours} (28,1) + low-pass scaling & \textbf{0.0701}$^{*}$ & \textbf{12.39}$^{*}$ & +.038\phantom{$^{*}$} & \textbf{0.046}$^{*}$ \\
\midrule
\rowcolor{gray!14}\multicolumn{5}{@{}l}{\emph{cost 16 (single-flow Euler steps)}} \\
single flow & 0.0727$^{*}$ & 13.08$^{*}$ & +.036\phantom{$^{*}$} & 0.030$^{*}$ \\
pixel-domain flow 600 h & 0.0735$^{*}$ & 12.79$^{*}$ & \textbf{+.042}\phantom{$^{*}$} & 0.036\phantom{$^{*}$} \\
single flow, reflowed \cite{liu2023rectified} & 0.0737$^{*}$ & 12.89$^{*}$ & +.038\phantom{$^{*}$} & 0.032$^{*}$ \\
single flow, reflowed + band scaling & 0.0716$^{*}$ & 12.45$^{*}$ & +.040$^{*}$ & 0.040$^{*}$ \\
single flow + churn \cite{karras2022edm} & 0.0830$^{*}$ & 13.02$^{*}$ & +.037\phantom{$^{*}$} & 0.031$^{*}$ \\
single flow + repulsion \cite{corso2024particle} & 0.0905$^{*}$ & 12.53$^{*}$ & +.039\phantom{$^{*}$} & 0.039$^{*}$ \\
single flow + band scaling & 0.0707\phantom{$^{*}$} & 12.65\phantom{$^{*}$} & +.039\phantom{$^{*}$} & 0.038\phantom{$^{*}$} \\
single flow + low-pass scaling & 0.0705$^{*}$ & 12.59$^{*}$ & +.038\phantom{$^{*}$} & 0.038$^{*}$ \\
\rowcolor{blue!9}\textbf{ours} (32,8) & 0.0703\phantom{$^{*}$} & 12.49$^{*}$ & +.039\phantom{$^{*}$} & 0.046$^{*}$ \\
\rowcolor{blue!9}\textbf{ours} (32,8) + low-pass scaling & \textbf{0.0698}$^{*}$ & \textbf{12.35}$^{*}$ & +.037\phantom{$^{*}$} & \textbf{0.047}$^{*}$ \\
\bottomrule
\end{tabular}}

\end{table}

\paragraph{Data and models.}
We used hourly IMERG-Late ($0.1^\circ$) as input and MRMS (1\,km)
\cite{zhang2016mrms} as the target on a $10{:}1$ grid over CONUS for 2021--2024. We held out 2024 as the test year and
four months of 2021--2023 for validation. We set $T(y)=\log(1+y)$ for rain in mm/h and
scored on the radar quality mask. The deterministic predictor used a local implicit decoder \cite{chen2021liif}
with a $3\times3$ coarse-cell context. All flows used db2 wavelets \cite{daubechies1992ten}, $256^2$ training tiles, an
8.5\,M-parameter FiLM U-Net family, and matched losses, optimizer and training budgets
($4\times10^4$ updates, batch 32). We trained both recursive flows on the same tile pool with
$\sigma_{\mathrm{cond}}=0.05$. Main models used 2000 training hours; allocation ablations used a matched 600-hour subset.

\paragraph{Baselines.}
The single flow generates all residual bands with one flow at 4, 8 or 16 Euler steps; the pixel-domain flow does the same without the wavelet transform (trained on the 600-hour
subset). Reflow \cite{liu2023rectified} is the single flow distilled once on its own sixteen-step
samples. Churn \cite{karras2022edm} and repulsion \cite{corso2024particle} add noise or
diversity at sampling time. Band scaling applies the closed-form all-band gain of Sec.~\ref{sec:spectral}; low-pass
scaling applies the validation-selected gain. No low-pass gain passed the guard for the
single flow or reflow below cost~sixteen.

\paragraph{Allocation and calibration settings.}
The fine and coarse grid sides were
$h_f=128$ and $h_c=64$, and the low-pass $L$ and details $D$ were the level-one db2 bands; $d_L$ denotes the low-pass deficit. A coarse step therefore processes a quarter of the coefficients of a fine step and, with the
convolutional U-Net, costs about a quarter as much, so we set $c_f=1$ and $c_c=1/4$. We
write an allocation as $(N_c,N_f)$, the numbers of coarse and fine steps. Its cost is
$N_c/4+N_f$ in units of single-flow Euler steps; for example, $(16,4)$ and $(28,1)$ both
cost eight.
We used budgets $B\in\{4,8,16\}$. We set $\Omega=2$--$6$\,km
and $[P_{\min},P_{\max}]=[0.8,1.25]$. The coarse-heavy candidates were $(12,1)$ and
$(28,1)$ at costs four and eight; equal-compute allocation at eight was $(16,4)$.
At cost sixteen, $(48,4)$ exceeded the spectral limit on validation, so we retained
$(32,8)$. The selected low-pass
gains were 1.50, 1.09 and 1.06, respectively. The closed-form low-pass gain 1.36 at $(28,1)$ raised the power ratio from 0.7 to 4.7.

\paragraph{Scoring.}
We evaluated the primary training run on 48 month-stratified hours of 2024 tiled into 1056
windows of $1024$\,km, with sixteen members, and repeated key comparisons on a second, independent set of 48 hours. We scored fair CRPS
\cite{gneiting2007proper,ferro2014fair}, conditional CRPS on observed rain $\ge10$\,mm/h,
Brier skill score (BSS) and member-wise critical success index (CSI) at 10\,mm/h, and the power ratio $P_{2\text{--}6}$. CRPS averaged tiles equally; CSI and Brier scores pooled sufficient statistics. Within each cost tier, we compared every row with the band-scaled single flow (single flow + band scaling in
Table~\ref{tab:cost}) by paired block bootstraps over hours (2000 resamples); asterisks mark differences whose 95\% interval excludes zero; zero-crossing intervals do
not establish equivalence. The Murphy components
of the Brier score \cite{murphy1973brier} served as calibration~diagnostics.

\paragraph{Sampling time.}
We measured residual sampling times on an NVIDIA RTX PRO 6000 GPU. We used FP32, sixteen members per $1024^2$
window and six repetitions on three validation windows, timing noise, transfers, tiling
and reconstruction only.
With member batching and a shared 250-tile cap, median times were \EBatchAllocated\,s for $(28,1)$ and
\EBatchFlatSixteen\,s for the sixteen-step single flow, close to the 0.5 that the proxy predicts for cost eight versus sixteen fine~steps.

\paragraph{Results overview.}
Table~\ref{tab:cost} shows the test scores of all methods at matched cost, grouped by budget. Table~\ref{tab:alloc} shows the allocation sweep of the 600-hour model. Fig.~\ref{fig:allocation} shows the validation CRPS along that sweep and the fixed-model seed
replication at cost~eight.

\subsection{Evaluating step allocation}

\begin{figure}[t]
\centering
\includegraphics[width=\columnwidth]{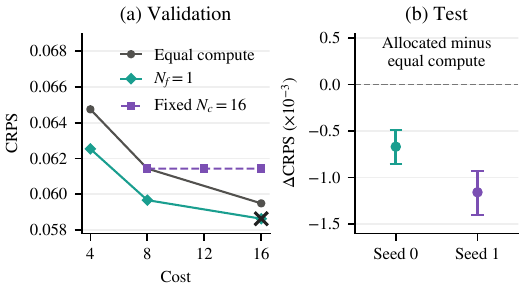}
\caption{(a) Validation CRPS of a 600-hour model; $\times$ marks the rejected $(60,1)$. (b) Test CRPS of
$(28,1)$ minus $(16,4)$ for two 2000-hour seeds at cost eight, with 95\% hour-bootstrap~intervals.}
\label{fig:allocation}
\end{figure}

\paragraph{Fixed-model control.}
Changing only the step allocation improved the uncalibrated ensemble.
At cost eight, moving from equal compute $(16,4)$ to $(28,1)$ on the same
2000-hour checkpoint lowered test CRPS from 0.0715 to 0.0708 and improved CSI
(Table~\ref{tab:cost}). For CRPS and CSI, the paired 95\% intervals of the difference excluded zero; the
Brier-skill difference was inconclusive (Fig.~\ref{fig:allocation}(b)).

With the second seed, the same allocation change again improved CRPS and CSI,
with paired intervals excluding zero. Heavy-rain CRPS and Brier skill also improved
in point estimates. Both comparisons used raw ensembles, so these gains did not depend on spread
correction.

\paragraph{Allocation sweep.}
The completed allocation sweep showed little benefit from extra fine steps
(Fig.~\ref{fig:allocation}(a), Table~\ref{tab:alloc}). With $N_c=16$, raising $N_f$ from 4 to 12 left validation and test CRPS unchanged at the
displayed precision.
At cost eight, shifting computation toward the coarse flow improved both validation
and test CRPS throughout the sweep. The low-pass deficit decreased with coarse steps,
while detail deficits changed little with extra fine steps. Every retained raw allocation was still under-dispersed at 2--6\,km ($P_{2\text{--}6}$ below 0.8,
Table~\ref{tab:alloc}), which the low-pass gain of Sec.~\ref{sec:spectral} had to restore. Adding a third level did not help: a three-level model with a third flow on the level-two
low-pass band, allocated by the same procedure, scored worse at cost eight both without
and with low-pass scaling, so we kept two~levels.

\paragraph{Rejected allocations.}
At cost sixteen, $(60,1)$ achieved the lowest validation CRPS but exceeded the spectral
limit and over-forecast heavy rain ($\times$ in Fig.~\ref{fig:allocation}(a), Table~\ref{tab:alloc});
its test Brier skill was also below the retained $(32,8)$.
For the 2000-hour model, $(48,4)$ exceeded the limit on validation and showed the same
tradeoff: better test CRPS but worse reliability and Brier skill than $(32,8)$.

\begin{table}[t]
\centering\footnotesize
\caption{Allocation sweep of the 600-hour model; Cost is $N_c/4+N_f$. Bold marks the best
score per cost, shading the retained~allocation.}
\label{tab:alloc}
\vspace{0.0pt}%
\setlength{\tabcolsep}{2pt}
\renewcommand{\arraystretch}{1.30}
\setlength{\aboverulesep}{0pt}\setlength{\belowrulesep}{0pt}
\begin{tabularx}{\columnwidth}{r>{\raggedright\arraybackslash}X*{5}{>{\raggedleft\arraybackslash}X}}
\toprule
 & & \multicolumn{3}{c}{Validation} & \multicolumn{2}{c}{Test} \\
\cmidrule(lr){3-5}\cmidrule(lr){6-7}
Cost & $(N_c,N_f)$ & $d_L$ & $P_{2\text{--}6}$ & CRPS$\downarrow$ & CRPS$\downarrow$ & BSS$\uparrow$ \\
\midrule
\rowcolor{blue!9}4 & (12,1) & 3.06 & 0.09 & \textbf{0.0625} & \textbf{0.0769} & \textbf{+.027} \\
\midrule
8 & (4,7) & 9.48 & 0.02 & 0.0692 & 0.0856 & +.008 \\
 & (8,6) & 4.52 & 0.03 & 0.0648 & 0.0795 & +.018 \\
 & (16,4) & 2.43 & 0.21 & 0.0614 & 0.0759 & +.033 \\
 & (24,2) & 1.89 & 0.47 & 0.0601 & 0.0747 & +.037 \\
\rowcolor{blue!9} & (28,1) & 1.75 & 0.59 & \textbf{0.0597} & \textbf{0.0744} & \textbf{+.038} \\
\midrule
16 & (8,14) & 4.52 & 0.03 & 0.0648 & 0.0795 & +.018 \\
 & (16,12) & 2.43 & 0.23 & 0.0614 & 0.0759 & +.033 \\
\rowcolor{blue!9} & (32,8) & 1.65 & 0.77 & 0.0595 & 0.0743 & \textbf{+.037} \\
 & (48,4) & 1.44 & 1.13 & 0.0589 & 0.0739 & +.034 \\
 & (60,1) & 1.37 & 1.29 & \textbf{0.0586} & \textbf{0.0737} & +.032 \\
\bottomrule
\end{tabularx}

\end{table}

\subsection{Predictive quality}

At cost eight, the recursive flow improved all four primary scores over the eight- and
sixteen-step single flows (Table~\ref{tab:cost}); against sixteen steps, CRPS fell from 0.0727
to 0.0708 with \ESpeedRange$\times$ faster~sampling.

The band-scaled single flow was a stronger baseline. At cost eight, it matched
the raw recursive flow in rounded CRPS and Brier skill, but had worse heavy-rain CRPS
and CSI. With low-pass correction, the recursive flow had the lowest CRPS at costs eight and
sixteen (Table~\ref{tab:cost}).
Over both sets of 48 test hours at cost eight, its paired 95\% CRPS interval against the band-scaled single flow was $[-0.0015,-0.0007]$;
CSI improved, and Brier skill remained~\mbox{inconclusive}.

\subsection{Calibration tradeoffs and robustness}

\paragraph{Cost four.}
At cost four, low-pass correction improved overall probabilistic accuracy, with a
heavy-rain tradeoff. It lowered recursive CRPS from 0.0727 to 0.0695, outperforming the band-scaled single flow and reflow with paired 95\% intervals excluding
zero (Table~\ref{tab:cost}). CSI improved against the single flow, while heavy-rain CRPS worsened against reflow;
Brier skill and the reflow CSI difference were~inconclusive.

\paragraph{Band scaling.}
All-band spread scaling reduced CRPS further than low-pass correction but worsened Brier
skill and reliability, trading lower CRPS for poorer calibration.

\paragraph{Robustness.}
On the second set of 48 test hours, the calibrated cost-four recursive model again
outperformed both band-scaled baselines in CRPS; over all 96 hours, its paired 95\% CRPS
interval against the band-scaled single flow was $[-0.0039,-0.0031]$, with Brier skill
inconclusive. A second training seed kept the CRPS
gain over the band-scaled single flow but not the Brier-skill and CSI gains, and its
heavy-rain CRPS was~worse.

\section{Conclusion}
\label{sec:concl}

We showed that a coarse-heavy step allocation, predicted from the band-wise
spread--error deficit, improved CRPS and CSI across training seeds at fixed cost.
The uncalibrated model outperformed the sixteen-step single flow while sampling
\ESpeedRange$\times$ faster. Low-pass scaling further improved probabilistic accuracy.

Evaluation was limited to precipitation, and heavy-rain skill varied across seeds.
The procedure is widely applicable to multiscale conditional generation, including image
super-resolution and atmospheric downscaling. We plan to evaluate these applications and
to improve heavy-rain calibration.\label{sec:body-end}

\clearpage
\bibliographystyle{IEEEbib}
\bibliography{refs}

\end{document}